\pdfoutput=1

\documentclass[11pt]{article}

\usepackage[preprint]{acl}
\usepackage{float}
\usepackage{tcolorbox}
\usepackage{enumitem}
\usepackage{amssymb}

\usepackage{times}
\usepackage{latexsym}
\usepackage{amsmath, amssymb}  
\usepackage{booktabs}  
\usepackage{multirow}  
\usepackage{array}     
\usepackage{tabularx}  

\usepackage[T1]{fontenc}

\usepackage[utf8]{inputenc}

\usepackage{microtype}

\usepackage{inconsolata}

\usepackage{graphicx}

\usepackage{booktabs}
\usepackage{colortbl}
\usepackage{amsmath}

\definecolor{Gray}{gray}{0.2}
\definecolor{lightgray}{gray}{0.92}
\definecolor{OurColor}{rgb}{0.886, 0.941, 0.851}

\title{MathVis-Fine: Aligning Visual Supervision with Necessity via Progressive Dependency-Guided Training for Multimodal Mathematical Reasoning
}

\author{
\textbf{Wanshi Xu}\textsuperscript{1,*},
\textbf{Haokun Zhao}\textsuperscript{2,*},
\textbf{Haidong Yuan}\textsuperscript{3},
\textbf{Songjun Cao}\textsuperscript{4},
\textbf{Long Ma}\textsuperscript{4,$\dagger$}
\\[6pt]
\textsuperscript{1} School of ECE, Peking University \\
\textsuperscript{2} College of Computer Science and Artificial Intelligence, Fudan University \\
\textsuperscript{3} School of Software and Microelectronics, Peking University \\
\textsuperscript{4} Tencent Youtu Lab \\[4pt]
\ttfamily \{xwanshi, oseast\}@stu.pku.edu.cn \\
\ttfamily hkzhao23@m.fudan.edu.cn,
\ttfamily \{songjuncao, malonema\}@tencent.com
\thanks{$*$ These authors contributed equally to this work. $\dagger$ Corresponding author.}
}

\begin{document}
\maketitle
\begin{abstract}
Chain-of-Thought (CoT) reasoning has extended from purely linguistic domains to multimodal scenarios; however, existing approaches often treat visual inputs as homogeneous or auxiliary signals, failing to capture the intricate and sample-specific dependencies between text and images in mathematical problem-solving. This gives rise to two core issues: first, the supervisory signals for visual content are generalized and coarse-grained, lacking adaptation to the actual necessity of visual information in each sample; second, training feedback becomes inaccurate when visual rewards are uniformly applied without distinguishing the complementary relationships among inputs. These limitations hinder models from achieving precise multimodal reasoning.
In this work, we propose a framework for modeling fine-grained visual dependencies in mathematical reasoning. We first construct the MathVis-Fine dataset, augmenting fine-grained visual annotations with visual dependency ratings. Building upon this dataset, we introduce a two-stage progressive visual enhancement training paradigm that balances answer correctness rewards and visual grounding rewards according to the intrinsic visual dependency level of each sample, thereby mitigating reward bias and improving supervision accuracy. Extensive experiments demonstrate that MathVis-Fine framework effectively enhances visual perception progressively based on visual dependency, offering a more precise training framework for multimodal mathematical reasoning.\footnote{We will release the dataset upon acceptance.}
\end{abstract}

\section{Introduction}

In recent years, the application of Multimodal Large Language Models (MLLMs)~\cite{openai2023gpt4v,liu2023llava} to mathematical reasoning tasks has achieved significant progress. Such tasks require models to process and integrate textual and visual information to solve complex problems. Although traditional Large Language Models (LLMs)~\cite{touvron2023llama,achiam2023gpt,yang2025qwen3} have demonstrated strong reasoning capabilities in purely textual domains, extending these capabilities to multimodal scenarios, particularly tasks involving mathematical charts, geometric figures, or symbolic visual representations, remains a challenging frontier.

The visual components in mathematical problems introduce unique difficulties~\cite{an2025empowering}. First, mathematical images often contain precise geometric relationships, symbolic annotations, and spatial configurations that general-purpose visual encoders or simple bounding-box annotations cannot adequately capture. Second, existing approaches tend to adopt a uniform visual processing pipeline, overlooking the actual degree of visual dependency in each problem. Some problems may rely heavily on visual information (e.g., geometric proofs), while others can be solved primarily through textual logic. Such oversimplification leads to two key issues: (1) insufficient visual grounding when images are critical, and (2) unnecessary computational overhead and potential noise when images are less relevant.

Current strategies to enhance the visual perception of mathematical MLLMs~\cite{xiao2025advancing,wang2025papo} typically involve strengthening visual attention mechanisms, introducing additional visual supervision and rewards, integrating external visual tools, or increasing the granularity of visual annotations. However, these methods share a common limitation: they treat all samples as having the same degree of visual dependency. In reality, there is significant variation in the reliance on visual information across different mathematical domains and even among different problem types. Ignoring this variability results in misaligned supervisory signals: excessive penalization of visual errors in text-dominant problems and insufficient emphasis on visual accuracy in visually critical problems.

To address these limitations, we propose the MathVis-Fine framework, which explicitly models and adapts to the varying visual dependencies in multimodal mathematical reasoning. Our core insight is that effective multimodal reasoning requires not only enhanced visual perception but also visual perception that is adapted to the specific visual needs of each problem. We make the following three key contributions:

\begin{itemize}
    \item MathVis-Fine Dataset: We construct a dataset comprising approximately 5.4 thousand mathematical problems, each annotated with fine-grained visual dependency ratings and step-level alignments between textual reasoning phrases and corresponding visual regions.
    \item Two-Stage Visual-Dependency Guided Training Pipeline: We develop a progressive training strategy that begins with cold-start SFT to enhance visual perception on highly dependent samples, and finally employs a visual-dependent reward mechanism during the reinforcement learning stage.
    \item Multi-Dimensional Visual Reward Mechanism: We introduce a multi-dimensional visual reward during the GRPO stage, effectively assessing the accuracy of visual region retrieval and visual content recognition, thereby enabling more precise and efficient feedback for visual perception.
\end{itemize}

Our experiments demonstrate that MathVis-Fine significantly outperforms existing methods across multiple multimodal mathematical benchmarks, validating the importance of modeling varying visual dependencies in mathematical reasoning.

\section{Related Work}

 \paragraph{MLLMs for Mathematics.}
In recent years, Multimodal Large Language Models (MLLMs)~\cite{openai2023gpt4v,liu2023llava,Bai2023QwenVLAF,jiang2024mmsearch} have demonstrated remarkable proficiency across diverse vision-language tasks. Consequently, a variety of specialized methodologies~\cite{gao2023g, zhang2024mavismathematicalvisualinstruction, huang2024autogeo, deng2024r, luo2025ursa, shi2024math, peng2024multimath} have emerged to bolster visual mathematical reasoning capabilities. For instance, approaches such as G-LLaVA~\cite{gao2023g} and Math-LLaVA~\cite{shi2024math} employ dataset augmentation strategies to expand data coverage, thereby adapting models to specialized mathematical tasks. Notably, MAVIS~\cite{zhang2024mavismathematicalvisualinstruction} introduces a fully automated data generation engine to curate large-scale mathematical visual datasets. It adopts a four-stage training pipeline: initially training a specialized vision encoder, followed by vision-language alignment, instruction tuning, and finally enhancing CoT reasoning via Direct Preference Optimization (DPO).
In the realm of reinforcement learning, MM-Eureka~\cite{meng2025mmeurekaexploringfrontiersmultimodal} extends Reinforcement Learning with Verifiable Rewards (RLVR) to mathematical reasoning tasks without cold-start initialization, achieving substantial improvements in multimodal reasoning. Furthermore, Vision-R1~\cite{huang2025vision} employs a training paradigm consisting of cold-start fine-tuning on long CoT data followed by large-scale RL, attaining state-of-the-art performance across multiple multimodal mathematical benchmarks.

\paragraph{Visual Chain-of-Thought.}
Capitalizing on advancements in visual reasoning tasks~\cite{lu2024mathvista,yue2024mmmu,jiang2025mmecotbenchmarkingchainofthoughtlarge}, Visual Chain-of-Thought (Visual CoT) has established itself as an effective paradigm for both image generation and comprehension~\cite{guo2025can, jiang2025t2i, tong2025delving, zhuo2025reflection,o1,yao2024mulberry, qwen_qvq_72b_preview}. Early iterations, such as Visual CoT~\cite{shao2024visual} and Chain-of-Spot~\cite{liu2024chain}, propose cropping highly attended image regions and integrating them into the chain-of-thought process. Despite demonstrating promising performance, these methods are often constrained by rigid image cropping heuristics or a dependency on external tools. In contrast, MINT-CoT~\cite{chen2025mint} enhances the grounding of visual information within the Visual CoT by introducing explicit retrieval targets during training. This approach refines the model's perception and attentional focus on fine-grained visual details essential for mathematical reasoning.

\paragraph{Perception Alignment in Reinforcement Learning.}
Although RLVR has driven significant progress in textual reasoning~\cite{wang2025papo, xiao2025advancing}, its direct application to the multimodal domain encounters a critical perception bottleneck. Recent studies indicate that standard RLVR often encourages models to bypass visual perception, leading to hallucinated correct answers derived from textual biases rather than valid visual evidence~\cite{an2025empowering}.
To address this, recent works propose incorporating perception-aware signals directly into the optimization process. Perception-R1~\cite{xiao2025advancing} introduces an explicit \textit{Visual Perception Reward}. By extracting atomic visual facts (e.g., geometric relations) from correct reasoning trajectories and employing a judge model to verify their presence in the generated rationale, it enforces a tighter alignment between visual input and textual output.
Conversely, PAPO~\cite{wang2025papo} proposes an implicit supervision mechanism via visual augmentation. It designs an implicit perception loss that penalizes the model if it generates high-confidence answers without relying on valid visual features. However, in complex mathematical reasoning, visual signals are highly structured and fine-grained. Uniform enhancement strategies implemented through global reward signals or image-level augmentation fail to distinguish the inherent heterogeneity of visual information.

\section{Methodology}

\label{sec:overview}
\begin{figure*}[t]
    \centering
    \includegraphics[width=1\linewidth]{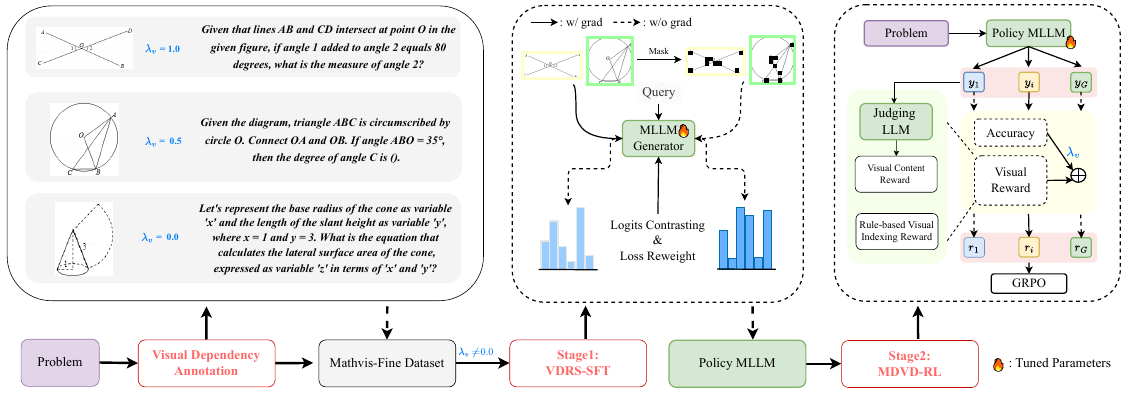}
    \caption{\textbf{Overview of the framework}, which begins by constructing a dataset with fine-grained visual dependency annotations. \textbf{Stage 1} employs a Retrieval-Perception Synergy strategy during supervised fine-tuning (SFT) to enhance visual perception. \textbf{Stage 2} utilizes Multi-Dimensional Visual-Dependent Reinforcement Learning (MDVD-RL). By integrating the two visual rewards and leveraging the dependency score ($\lambda_v$) as a gating factor, this stage further elevates fine-grained multimodal reasoning capabilities via GRPO.}
    \label{fig:overview}
\end{figure*}

\subsection{MathVis-Fine Dataset}
\label{sec:mathvis}
To empower the framework with heterogeneous visual perception, we develop an enhanced data generation pipeline that produces mathematical visual-interleaved samples with both fine-grained token-level alignment annotations and visual dependency scores. This pipeline yields a dataset of 5.4K high-quality training samples for SFT and RL.

\noindent \textbf{Data Generation and Structure}
We construct our dataset based on mathematical problems sourced from the MINT-CoT dataset~\cite{chen2025mint}, which provides high-quality reasoning chains annotated with fine-grained visual patch index alignment. The motivation for introducing the visual dependency annotation $\lambda_v$ stems from a key limitation of prior methods: treating all multimodal problems as having uniform visual importance. By explicitly quantifying the degree of visual necessity, our dataset enables training pipelines to: (i) focus computational resources on samples where visual information is critical during training; (ii) design loss functions and reward mechanisms weighted by $\lambda_v$, ensuring that visual grounding is emphasized proportionally to its actual relevance; and (iii) facilitate more nuanced evaluation of model performance across problems with varying visual demands.

\noindent \textbf{Visual Dependency Annotation:} This annotation is implemented via a rule-based protocol followed by manual sampling inspection. The annotation process is performed automatically by providing the model with the problem text, the associated image, and a structured prompt that guides the evaluation based on the defined criteria. The final output categorizes visual dependency into three levels:

\begin{itemize}
    \item $\boldsymbol{\lambda_v = 1.0}$ (\textbf{High}): The image contains core and indispensable information required to solve the problem (e.g., unstated geometric relationships, chart data). The problem cannot be solved without the visual input.
    \item $\boldsymbol{\lambda_v = 0.5}$ (\textbf{Medium}): The image provides important contextual or clarifying information that complements the text. The problem is theoretically solvable using text alone, but would be significantly more difficult or ambiguous without the image.
    \item $\boldsymbol{\lambda_v = 0.0}$ (\textbf{Low}): The image is redundant or purely decorative. All necessary information is fully and accurately described in the text.
\end{itemize}

Each finalized sample in our dataset constitutes a structured input-output pair. The input includes the original mathematical question $q$ (text) and its corresponding diagram, chart, or graphical representation $I$ (image). The output comprises: (1) a \textit{visual-interleaved chain of thought}, i.e., a step-by-step reasoning sequence where textual references to visual elements are annotated with corresponding spatial grid indices $\mathcal{I}_{\text{gt}}$, for instance, “Calculate the length of line AB (${index: 15, 16}$); (2) a \textit{visual dependency score} $\lambda_v$ that quantifies the degree to which solving the problem relies on visual information, taking a discrete value of $0.0$, $0.5$, or $1.0$; and (3) the \textit{final answer} to the mathematical problem.

\subsection{Training Strategy}
\label{sec:training_strategy}

Building upon the previously introduced MathVis-Fine dataset, we propose a phased training strategy. This strategy begins with a model capable of interleaved chain-of-thought reasoning. Specifically, we employ the supervised fine-tuning (SFT) model from~\citet{chen2025mint} that completes interleaved token generation. We first strengthens visual perception through enhanced training on data with high visual dependency, and subsequently employs reinforcement learning to improve model performance and generalization further.

\paragraph{Stage 1: Visual Dependency Training with Retrieval-Perception Synergy (VDRS-SFT)}
\label{sec:stage1}

The initial model possesses the basic capability to perform explicit vision-text interleaved reasoning via an interleave token mechanism, aided by a Binary Cross-Entropy (BCE) loss associated with a visual retrieval module. The model output is formatted as:
\begin{multline}
\{ v^{(1)}, s^{(1)}, v^{(2)}, s^{(2)}, \dots, v^{(k)}, s^{(k)} \}, \\
\text{answer} \sim P_\theta(\cdot \mid I, q),
\end{multline}
where $v^{(i)}$ denotes the selected visual tokens and $s^{(i)}$ represents the textual reasoning steps.

Our core observation is that for problems with high visual dependency ($\lambda_v \neq 0.0$), accurate visual retrieval and the model's genuine reliance on visual information should be synergistically optimized and mutually reinforcing. Specifically, the more precise the retrieved visual evidence, the more significantly the model's reasoning performance should degrade when that visual information is partially occluded. This reflects that the model indeed depends on the visual content rather than textual priors. To this end, we design a unified Retrieval-Perception Synergy Loss, which integrates explicit retrieval supervision with implicit dependency verification into a coherent optimization objective.

\textbf{Base Supervision Objectives:} We employ the cross-entropy loss for textual reasoning ($\mathcal{L}_{\text{CE}}$) and the binary cross-entropy loss for visual retrieval ($\mathcal{L}_{\text{BCE}}$) as the foundational supervision:
\begin{equation}
\mathcal{L}_{\text{CE}} = -\sum_{t \in \mathbf{q}} \log P_\theta \big( y_t \mid y_{<t}, I, q \big),
\end{equation}

\begin{equation}
\begin{split}
\mathcal{L}_{\text{BCE}} &= - \sum_{i=1}^{N} \sum_{j=1}^{L} \Big[ X_{ij} \log \sigma(\alpha_{ij}) \\
&\quad + (1 - X_{ij}) \log \big(1 - \sigma(\alpha_{ij})\big) \Big].
\end{split}
\end{equation}
where \(N\) is the number of Interleaved Tokens in a batch, \(L\) is the length of input visual tokens, \(\sigma(\cdot)\) denotes the sigmoid function and \(\alpha\) is the similarity scores introduced in \cite{chen2025mint} with ground-truth labels \(X \in \{0,1\}\) from $\mathcal{I}_{\text{gt}}$.

\textbf{Retrieval-Perception Synergy Loss:} For samples with visual dependency ($\lambda_v \neq 0.0$), we introduce a joint loss function aimed at simultaneously optimizing retrieval accuracy and regularizing the model's behavioral consistency under visual masking conditions. Its formulation is as follows:
\begin{multline}
\mathcal{L}_{\text{Synergy}} = \mathcal{L}_{\text{BCE}} + \gamma \cdot \mathbb{E}_{I_{\text{mask}} \sim \mathcal{M}(I)} \\
\Big[ D_{\text{KL}} \big( \pi_\theta(\cdot \mid q, I) \, \| \, \pi_\theta(\cdot \mid q, I_{\text{mask}}) \big) \Big].
\end{multline}

Here, $I_{\text{mask}}$ is a version of the original image $I$ where the key visual regions identified via retrieval are masked. $\mathcal{M}$ denotes the masking operation distribution. $\pi_\theta(\cdot \mid q, I)$ and $\pi_\theta(\cdot \mid q, I_{\text{mask}})$ represent the model's output distributions conditioned on the complete image and the masked image, respectively. $\gamma$ is the synergy weighting coefficient.

Therefore, the complete training objective for high-dependency samples is:
\begin{equation}
\mathcal{L}_{\text{High}} = \mathcal{L}_{\text{CE}} + \mathcal{L}_{\text{Synergy}} 
\end{equation}

This ensures that in vision-critical tasks, the model learns not only to retrieve correct visual evidence but also to substantively depend on this evidence during reasoning, thereby achieving an intrinsic unification of explicit visual grounding and implicit visual dependency.

\paragraph{Stage 2: Multi-Dimensional Visual-Dependent Reinforcement Learning (MDVD-RL).}

While the supervised training in Stage 1 establishes a foundational capability for interleaved reasoning, it is limited by the static nature of ground-truth annotations. To enable the model to autonomously explore more flexible and effective visual token selection strategies guided by inference outcomes, we employ Reinforcement Learning. Specifically, we extend the Group Relative Policy Optimization (GRPO) framework~\cite{shao2024deepseekmathpushinglimitsmathematical} by introducing a multi-dimensional, dependency-aware reward mechanism.

For a given problem input $x$, we sample a group of $G$ outputs $\{y_1, y_2, \dots, y_G\}$ from the current policy $\pi_\theta$. We design three distinct reward components to evaluate these outputs:

\noindent \textbf{Visual Indexing Reward ($r_{\text{idx}}$).} 
To encourage precise localization of visual evidence, we evaluate the overlap between the model's retrieved visual indices ($\mathcal{I}_{\text{pred}}$) and the ground-truth indices ($\mathcal{I}_{\text{gt}}$). Following the intuition that high-quality retrieval requires both accuracy and coverage, we define the index reward as the product of Precision and Recall:
\begin{equation}
    r_{\text{idx}} = \underbrace{\frac{|\mathcal{I}_{\text{pred}} \cap \mathcal{I}_{\text{gt}}|}{|\mathcal{I}_{\text{pred}}| + \epsilon}}_{\text{Precision}} \times \underbrace{\frac{|\mathcal{I}_{\text{pred}} \cap \mathcal{I}_{\text{gt}}|}{|\mathcal{I}_{\text{gt}}| + \epsilon}}_{\text{Recall}},
\end{equation}
where $\epsilon$ is a small constant for numerical stability. This metric strictly penalizes both hallucinated tokens (low precision) and missed key regions (low recall).

\noindent \textbf{Visual Content Reward ($r_{\text{con}}$).} 
Index alignment alone does not guarantee semantic understanding. To assess whether the retrieved regions actually contain the necessary information, we utilize a frozen, high-performance Vision-Language Model as a judge, denoted as $\Phi_{\text{judge}}$. For each generated reasoning chain, the judge verifies if the visual content described in the ground truth has been correctly identified and interpreted, returning a binary score $s_k \in \{0, 1\}$ for each of the $K$ key visual attributes. The content reward is the average recognition rate:
\begin{equation}
    r_{\text{con}} = \frac{1}{K} \sum_{k=1}^{K} s_k.
\end{equation}

\noindent \textbf{Dependency-Adaptive Reward Fusion.} 
A core insight of our method is that visual supervision should be proportional to the problem's actual visual necessity. We utilize the visual dependency score $\lambda_v \in \{0.0, 0.5, 1.0\}$ (defined in Sec.~\ref{sec:mathvis}) as a gating factor. The total reward $r_j$ for the $j$-th sample is formulated as:
\begin{equation}
    r_j = r_{\text{ans}} + \eta \cdot \lambda_v \cdot \left( \frac{r_{\text{idx}} + r_{\text{con}}}{2} \right),
\end{equation}
where $r_{\text{ans}} \in \{0, 1\}$ indicates the correctness of the final answer, and $\eta$ is a hyperparameter balancing reasoning and visual grounding. This formulation ensures that for text-only problems ($\lambda_v=0$), the model optimizes solely for logical correctness, whereas for high-dependency tasks, it receives strong feedback on visual perception.

Finally, we compute the advantage $\hat{A}_j$ using group-relative normalization and optimize the policy via the GRPO objective:

\begin{equation}
\begin{split}
\mathcal{L}_{\text{GRPO}} &= - \mathbb{E}_{y \sim \pi_{\theta}} 
\bigg[ \frac{1}{G} \sum_{j=1}^G \Big( \frac{\pi_\theta(y_j)}{\pi_{\theta_{\text{old}}}(y_j)} \hat{A}_j \\
&\qquad - \beta D_{\text{KL}}[\pi_\theta \parallel \pi_{\text{ref}}] \Big) \bigg].
\end{split}
\end{equation}
Here, $\hat{A}_j = (r_j - \mu_{\mathbf{r}}) / \sigma_{\mathbf{r}}$ is the standardized advantage within the group, and the KL divergence term ensures the policy does not deviate excessively from the reference model $\pi_{\text{ref}}$ trained in Stage 1.

\section{Experiments}
\label{sec:bibtex}
\subsection{Datasets and Settings}
Following the settings of previous influential \cite{chen2025mint} work, we build on Qwen2-VL-7B~\cite{wang2024qwen2vlenhancingvisionlanguagemodels} and train our model with a combination of SFT and RL on the Mathvis-Fine dataset. All model parameters except the vision encoder are updated. For ease of comparison, we follow the testing and evaluation benchmark of MINT-CoT \cite{chen2025mint}. To enhance the assessment of fine-grained visual understanding capabilities, we introduce the HC-M3D \cite{hcm3d} evaluation benchmark.

\subsection{Implementation Details}

The architecture follows the MINT-CoT\cite{chen2025mint} configuration and is equipped with retrieval capabilities. We use Qwen-3-32B\cite{yang2025qwen3} as the judging model as the implement of Perception-R1\cite{xiao2025advancing}.
The training procedure consists of the following stages: Cold-Start Visual Perception SFT: We train on the MathVis-Fine dataset for 2 epochs, using a learning rate of $1 \times 10^{-6}$ and a batch size of 64. The synergy weight $\gamma$ is set to $0.2$. Multi-Dimensional Visual-Dependent Reinforcement Learning: We train on the MathVis-Fine dataset for 660 steps (2 epochs), utilizing a group size $G = 4$, a weighting factor $\eta = 0.9$, a learning rate of $1 \times 10^{-6}$, and a batch size of 16.
During training, all model parameters are unfrozen, except for the vision encoder.

\definecolor{bestclosed}{HTML}{E6F0FF} 
\definecolor{bestopen}{HTML}{E6FFE6}   

\begin{table*}[t]
\centering
\setlength{\tabcolsep}{15pt} 
\renewcommand{\arraystretch}{1} 

\resizebox{\linewidth}{!}{
    \begin{tabular}{l c c c c c c}
    \toprule
    \toprule
    \multirow{2}{*}{\textbf{Model}} & \multirow{2}{*}{\textbf{\#Params}} & \multicolumn{5}{c}{\textbf{MathVista-Math}} \\
    \cmidrule(lr){3-7}
     & & \textbf{All} & \textbf{GEO} & \textbf{ALG} & \textbf{GPS} & \textbf{TQA} \\
    \midrule

    \multicolumn{7}{c}{\textit{Closed-Source Models}} \\
    \midrule
    GPT-4o~\cite{openai2024gpt4ocard} & -- & 66.67 & 63.68 & 67.04 & 63.46 & \cellcolor{bestclosed}\textbf{77.42} \\
    Claude-3.5 Sonnet~\cite{AnthropicModelCA} & -- & \cellcolor{bestclosed}\textbf{67.41} & \cellcolor{bestclosed}\textbf{65.09} & \cellcolor{bestclosed}\textbf{67.79} & \cellcolor{bestclosed}\textbf{65.38} & 74.19 \\
    
    \midrule
    \multicolumn{7}{c}{\textit{Open-Source General MLLMs}} \\
    \midrule
    LLaVA-OneVision-Qwen2-7b-ov~\cite{li2024llava} & 7B & 67.04 & 69.34 & 67.04 & 69.71 & 58.06 \\
    InternVL2-8B~\cite{chen2024internvl} & 8B & 62.59 & 62.26 & 62.92 & 62.50 & 62.90 \\
    InternVL2-8B-MPO~\cite{wang2024enhancing} & 8B & 68.52 & 68.87 & 68.91 & 69.71 & 64.52 \\
    DeepSeek-VL2~\cite{wu2024deepseekvl2mixtureofexpertsvisionlanguagemodels} & 4.5B & 65.56 & 63.68 & 65.54 & 63.94 & 70.97 \\
    Qwen2.5-VL-7B-Instruct~\cite{bai2025qwen25vltechnicalreport} & 7B & 66.66 & 65.56 & 66.29 & 65.87 & 69.35 \\

    \midrule
    \multicolumn{7}{c}{\textit{Open-Source Reasoning MLLMs}} \\
    \midrule
    Open-R1-Multimodal~\cite{open-r1-multimodal} & 7B & 54.81 & 52.36 & 54.68 & 53.37 & 59.68 \\
    R1-VL-7B~\cite{zhang2025r1vllearningreasonmultimodal} & 7B & 69.63 & 68.87 & 69.66 & 69.71 & 69.35 \\
    Mulberry~\cite{yao2024mulberry} & 7B & 68.52 & 67.92 & 68.54 & 68.75 & 67.74 \\
    MM-Eureka~\cite{meng2025mm} & 7B & 72.59 & 71.22 & 72.66 & 72.60 & 72.58 \\
    MINT-CoT-7B~\cite{chen2025mint} & 7B & 73.70 & 74.53 & 73.78 & 75.00 & 69.35 \\

    \midrule
    \multicolumn{7}{c}{\textit{Our Method}} \\
    \midrule
    \textbf{MathVis-Fine (Ours)} & 7B & \cellcolor{bestopen}\textbf{77.26} & \cellcolor{bestopen}\textbf{77.85} & \cellcolor{bestopen}\textbf{77.62} & \cellcolor{bestopen}\textbf{78.41} & \cellcolor{bestopen}\textbf{72.90} \\
    
    \bottomrule
    \bottomrule
    \end{tabular}
}
\caption{\textbf{Quantitative comparison on MathVista (Mathematical Subset).} We compare our proposed \textbf{MathVis-Fine} against state-of-the-art closed-source and open-source MLLMs. Best results for \colorbox{bestclosed}{\textbf{Closed-Source}} and \colorbox{bestopen}{\textbf{Open-Source}} models are highlighted respectively.}
\label{tab:mathvista_results}
\end{table*}

\begin{table}[t]
\centering
\setlength{\tabcolsep}{1pt} 
\renewcommand{\arraystretch}{1.0} 

\definecolor{bestopen}{HTML}{E6FFE6}

\resizebox{\linewidth}{!}{
    \begin{tabular}{l c}
    \toprule
    \toprule
    \textbf{Model} & \textbf{GeoQA} \\
    \midrule
    
    Qwen2.5-VL-7B-Instruct~\cite{bai2025qwen25vltechnicalreport} & 43.50 \\
    Open-R1-Multimodal~\cite{open-r1-multimodal} & 48.67 \\
    Hint-GRPO~\cite{huang2025boostingmllmreasoningtextdebiased} & 55.31 \\
    R1-V~\cite{chen2025r1v} & 59.00 \\
    MINT-CoT-7B~\cite{chen2025mint} & \underline{64.72} \\
    
    \midrule
    
    \textbf{MathVis-Fine (Ours)} & \cellcolor{bestopen}\textbf{66.45} \\
    
    \bottomrule
    \bottomrule
    \end{tabular}
}
\caption{\textbf{Quantitative comparison on GeoQA.} We evaluate MathVis-Fine against the baseline and state-of-the-art models. \textbf{Bold} and \underline{underlined} indicate the best and second-best results, respectively.}
\label{tab:geoqa}
\end{table}

\begin{table}[t]
\centering
\setlength{\tabcolsep}{4pt} 
\renewcommand{\arraystretch}{1.1} 

\definecolor{bestopen}{HTML}{E6FFE6}

\resizebox{\linewidth}{!}{
    \begin{tabular}{l c}
    \toprule
    \toprule
    \textbf{Model} & \textbf{MMStar-Math} \\
    \midrule
    
    Qwen2.5-VL-7B~\cite{bai2025qwen25vltechnicalreport} & 66.8 \\
    InternVL2-8B~\cite{chen2024internvl} & 66.8 \\
    R1-VL-7B~\cite{r1vl} & 68.4 \\
    Mulberry-7B~\cite{yao2024mulberry} & 66.8 \\
    Open-R1-MM~\cite{open-r1-multimodal} & 59.2 \\
    MINT-CoT-7B~\cite{chen2025mint} & \underline{69.6} \\
    
    \midrule
    
    \textbf{MathVis-Fine (Ours)} & \cellcolor{bestopen}\textbf{71.0} \\
    
    \bottomrule
    \bottomrule
    \end{tabular}
}
\caption{\textbf{Combined results on the mathematical subset of MMStar.} We evaluate MathVis-Fine against the baseline and state-of-the-art models. \textbf{Bold} and \underline{underlined} indicate the best and second-best results, respectively.}
\label{tab:mmstar}
\end{table}

\begin{table*}[t]
\centering
\setlength{\tabcolsep}{15pt} 
\renewcommand{\arraystretch}{1} 

\resizebox{0.95\textwidth}{!}{%
\begin{tabular}{l|c|c|ccccc}
    \toprule
    \toprule
    \multirow{2}{*}{\textbf{Model Variant}} & 
    \multirow{2}{*}{\textbf{MMStar-Math}} &
    \multirow{2}{*}{\textbf{GeoQA}} &  \multicolumn{5}{c}{\textbf{MathVista-Math}}  \\
    \cmidrule{4-8}
    & & & \textbf{All} & \textbf{GEO} & \textbf{ALG} & \textbf{GPS} & \textbf{TQA} \\
    \midrule
    
    \rowcolor{gray!10} \multicolumn{8}{l}{\textit{Reference Model}} \\
    MINT-CoT-7B & 69.6 & 64.72 & 73.70 & 74.53 & 73.78 & 75.00 & 69.35 \\
    
    \midrule
    \rowcolor{gray!10} \multicolumn{8}{l}{\textit{Ablation Settings (Ours)}} \\

    w/o SFT Synergy Loss & 70.0 & 65.25 & 74.90 & 75.62 & 75.06 & 75.95 & 70.10 \\

    w/o Dependency Gating ($\lambda_v$) & 70.1 & 65.10 & 74.65 & 75.35 & 74.84 & 75.50 & 69.86 \\
    
    w/o Content Reward ($r_{\text{con}}$) & 70.0 & 66.15 & 76.53 & 77.16 & 76.80 & 77.52 & 71.84 \\

    w/o Index Reward ($r_{\text{idx}}$) & 70.1 & 65.90 & 76.22 & 76.84 & 76.45 & 77.26 & 71.52 \\
    
    \midrule
    \textbf{MathVis-Fine (Full)} & \cellcolor{bestopen}\textbf{71.0} & \cellcolor{bestopen}\textbf{66.45} & \cellcolor{bestopen}\textbf{77.26} & \cellcolor{bestopen}\textbf{77.85} & \cellcolor{bestopen}\textbf{77.62} & \cellcolor{bestopen}\textbf{78.41} & \cellcolor{bestopen}\textbf{72.90} \\

    \textit{$\Delta$ vs. MINT-CoT} & \textcolor{teal}{+1.4} & \textcolor{teal}{+1.73} & \textcolor{teal}{+3.56} & \textcolor{teal}{+3.32} & \textcolor{teal}{+3.84} & \textcolor{teal}{+3.41} & \textcolor{teal}{+3.55} \\
    
    \bottomrule
    \bottomrule
\end{tabular}
}
\caption{\textbf{Ablation study of MathVis-Fine components.} We evaluate the contribution of each module by removing it from the full pipeline and comparing against MINT-CoT. ``w/o SFT Synergy Loss'' indicates using standard SFT in Stage 1. ``w/o Dependency Gating'' implies applying visual rewards to all samples regardless of visual necessity, which introduces noise in text-heavy tasks.}
\label{tab:ablation_study}
\end{table*}

\definecolor{bestclosed}{HTML}{E6F0FF}
\definecolor{bestopen}{HTML}{E6FFE6}

\begin{table}[t]
\centering
\setlength{\tabcolsep}{5pt} 
\renewcommand{\arraystretch}{1.15} 

\resizebox{\linewidth}{!}{
    \begin{tabular}{l c c c c}
    \toprule
    \toprule
    \textbf{Model} & \textbf{ALL} $\uparrow$ & \textbf{DI} $\uparrow$ & \textbf{BC} $\uparrow$ & \textbf{AG} $\downarrow$ \\
    \midrule
    
    \multicolumn{5}{c}{\textit{Closed-Source Models}} \\
    \midrule
    GPT-4o~\cite{openai2024gpt4ocard} & \cellcolor{bestclosed}\textbf{49.0} & \cellcolor{bestclosed}\textbf{45.8} & \cellcolor{bestclosed}\textbf{19.1} & \cellcolor{bestclosed}\textbf{42.0} \\
    
    \midrule
    \multicolumn{5}{c}{\textit{Open-Source SOTA}} \\
    \midrule
    MultiMath-7B~\cite{peng2024multimath} & 49.2 & 44.8 & 16.6 & 56.9 \\
    InternVL2-8B~\cite{chen2024internvl} & 41.9 & 38.3 & 16.6 & \textbf{34.0} \\
    MINT-CoT-7B~\cite{chen2025mint} & \underline{48.2} & \underline{46.0} & \underline{18.5} & 38.5 \\
    
    \midrule
    \multicolumn{5}{c}{\textit{Our Method}} \\
    \midrule
    \textbf{MathVis-Fine (Ours)} & \cellcolor{bestopen}\textbf{50.6} & \cellcolor{bestopen}\textbf{47.2} & \cellcolor{bestopen}\textbf{20.5} & \cellcolor{bestopen}\textbf{34.8} \\
    
    
    \bottomrule
    \bottomrule
    \end{tabular}
}
\caption{\textbf{Evaluation results on the HC-M3D benchmark.} The $\uparrow$ and $\downarrow$ arrows indicate that higher or lower values are preferred, respectively. }
\label{tab:dataset_m3d}
\end{table}

\subsection{Main Results}
\label{sec:main_results}

\paragraph{Performance on General Mathematical Reasoning.}
Table~\ref{tab:mathvista_results} presents the comparative results on the MathVista benchmark. MathVis-Fine achieves a state-of-the-art accuracy of 77.26\% on the metric among open-source 7B models. This performance outperforms the recent strong reasoning model MINT-CoT by 3.56\%. A closer look at the sub-tasks reveals that our method excels particularly in categories requiring intensive visual interpretation. Specifically, in the Geometry (GEO) and GPS navigation subtasks, MathVis-Fine yields consistent gains of 3.32\% and 3.41\%, respectively.
 As shown in Table~\ref{tab:geoqa}, on the geometry-intensive GeoQA benchmark, MathVis-Fine achieves an accuracy of 66.45\%. This result not only surpasses the strong baseline MINT-CoT by 1.73\% but also outperforms the R1-V model by over 7\%. This substantial gain validates that our progressive training strategy effectively enhances the model's capability to interpret and utilize complex visual diagrams for logical reasoning.
 
\paragraph{Robustness and Fine-Grained Evaluation.}
We further evaluate the robustness of our model on MMStar-Math (Table~\ref{tab:mmstar}), where MathVis-Fine reaches 71.0\%, demonstrating consistent improvements over generalist models like InternVL2-8B. 
Moreover, we conduct a specialized evaluation on \textbf{HC-M3D}, a fine-grained mathematical benchmark designed to assess the precision of multimodal understanding and the rate of hallucinations. As detailed in Table~\ref{tab:dataset_m3d}, MathVis-Fine demonstrates a critical advantage: it achieves the best performance across almost all metrics while reducing the Attribute Generation (AG) error rate to 34.8. Since the AG metric measures the frequency of hallucinated visual attributes (where lower values indicate better grounding), this reduction compared to the other competitive baseline strongly supports our hypothesis. It indicates that the Visual Content Reward ($r_{\text{con}}$) effectively suppresses visual hallucinations by penalizing reasoning steps that reference non-existent visual features, ensuring that the generated reasoning is faithfully grounded in the image.

\subsection{Ablation Study}
\label{sec:ablation}
To investigate the specific contribution of each component within the MathVis-Fine framework, we conduct a comprehensive ablation study using MINT-CoT as the reference point. The results are summarized in Table~\ref{tab:ablation_study}.

\noindent\textbf{Impact of Synergy Loss (Stage 1).}
Replacing the Retrieval-Perception Synergy Loss with a standard BCE loss (``w/o SFT Synergy Loss'') leads to a performance drop of approximately 2.1\% on MathVista compared to the full model. Standard SFT typically treats visual retrieval and answer generation as separate objectives. Our results suggest that without the Synergy Loss, the model fails to establish a strong causal link between the retrieved visual tokens and the subsequent reasoning steps. The Synergy Loss forces the model to not only attend to visual tokens but to genuinely rely on them, as the reasoning loss is coupled with the retrieval quality under masking perturbations.

\noindent\textbf{Necessity of Dependency Gating ($\lambda_v$).}
A core design of our framework is the adaptive reward fusion based on the visual dependency score $\lambda_v$. The setting ``w/o Dependency Gating'' applies visual rewards to all samples uniformly, regardless of their actual visual need. As shown in Table~\ref{tab:ablation_study}, this results in a notable performance degradation (e.g., -2.46\% on MathVista All). We attribute this to the introduction of visual noise in text-dominant problems ($\lambda_v=0$). When the model is forced to find visual evidence for problems that are solvable by text alone, it may hallucinate visual connections or become distracted from the pure logical reasoning path. This confirms that selective visual supervision is superior to universal visual supervision.

\noindent\textbf{Complementarity of Visual Rewards.}
We also analyze the impact of the two distinct visual rewards. Ablating the Content Reward ($r_{\text{con}}$) causes a decline in TQA scores, suggesting that while the model may look at the correct region, it might misinterpret the semantic meaning without content-aware feedback. Conversely, removing the Index Reward ($r_{\text{idx}}$) leads to imprecise grounding, particularly harming performance on Geometry tasks where spatial precision is paramount. The full model achieves the best balance, demonstrating that spatial localization and semantic verification are mutually reinforcing components.

\subsection{Analysis on Perception-Accuracy Correlation}
\label{sec:correlation_analysis}

To further validate the premise of our visual dependency modeling, we investigate the relationship between visual perception quality and final reasoning accuracy across different problem types. We sample 100 samples for each dependency level and calculate the Pearson correlation coefficient between the two visual rewards and the binary correctness of the final answer on the MINT-CoT subset.
\begin{figure}[t]
    \centering
    \includegraphics[width=\linewidth]{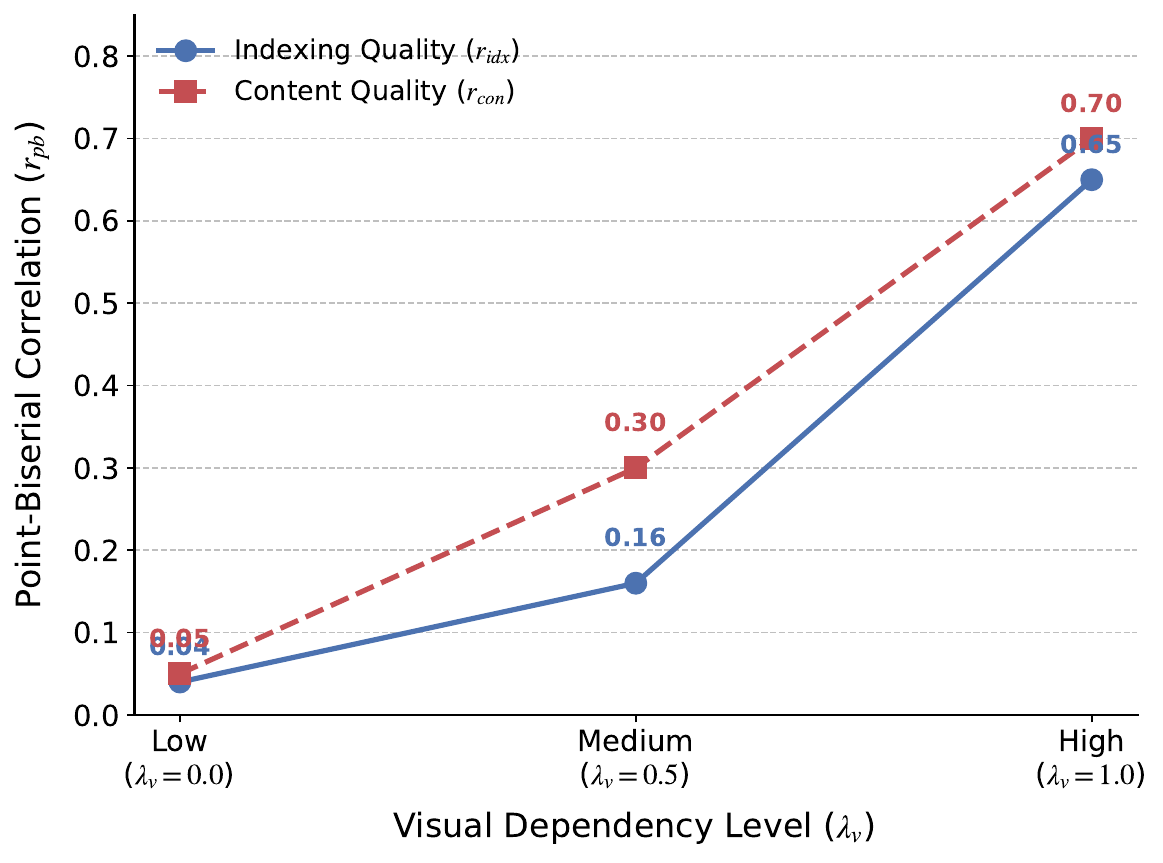} 
    \caption{\textbf{Pearson correlation coefficient between Visual Retrieval Recall and Answer Correctness across different visual dependency levels ($\lambda_v$), when $p < 0.05$.} The correlation significantly increases as the visual dependency of the problem rises, validating our strategy to weight visual rewards based on $\lambda_v$.}
    \label{fig:correlation}
\end{figure}

Figure~\ref{fig:correlation} illustrates these correlation scores grouped by the ground-truth visual dependency levels ($\lambda_v$).
    \noindent \textbf{Low Dependency ($\lambda_v=0.0$):} The correlation is near zero ($r \approx 0.05$). This indicates that for text-dominant problems, the model can answer correctly even if the visual retrieval is noisy or irrelevant. In these cases, enforcing strong visual rewards would be misleading, justifying our gating mechanism ($\lambda_v$ term in Eq. 8) which suppresses visual feedback for these samples.
    \noindent\textbf{High Dependency ($\lambda_v=1.0$):} We observe a strong positive correlation ($r > 0.65$). This confirms that for geometry and chart-based problems, accurate identification of visual elements is critical for successful reasoning. The high correlation supports our design of intensifying visual rewards (Index and Content) specifically for these high-dependency samples, as improvements in retrieval directly translate to gains in reasoning accuracy.

This analysis empirically proves that visual grounding is not universally equally important. By aligning the strength of visual supervision with this intrinsic correlation, MathVis-Fine effectively allocates optimization focus where it matters most.
\section{Conclusion}

In this work, we identified and addressed the limitations of existing multimodal CoT approaches, specifically their coarse-grained treatment of visual information and uniform reward mechanisms. We introduced MathVis-Fine, a high-quality dataset with fine-grained visual dependency annotations, and proposed a novel multi-stage training framework. By incorporating a Retrieval-Perception Synergy Loss in the supervised stage and a Dependency-Adaptive Reward mechanism in the reinforcement learning stage, our method effectively aligns visual supervision with the intrinsic visual necessity of each problem.
Extensive experiments demonstrate that our approach significantly outperforms state-of-the-art open-source models and effectively reduces visual hallucinations. Our findings highlight the importance of adaptive visual grounding in mathematical reasoning, paving the way for more precise and efficient Multimodal Large Language Models.
\section*{Limitations}

While MathVis-Fine successfully aligns visual supervision with the general degree of visual dependency ($\lambda_v$), our current approach to visual perception enhancement allows for deeper exploration regarding the \textit{granularity} of visual features. 
Specifically, mathematical problems exhibit varying sensitivities to different visual attributes. For instance, geometry problems often demand high sensitivity to spatial attributes, e.g., topology, relative positions. In contrast, data analysis tasks may depend more critically on discriminative precision. 
Our current reward mechanism optimizes for general retrieval accuracy without explicitly disentangling these attribute-specific requirements. 
Consequently, future work requires the integration of stronger vision-specialized prior models to perform visual preference distillation. By transferring knowledge regarding specific attribute sensitivities, we aim to achieve a more adaptive visual perception framework that dynamically adjusts its attention focus based on the distinct perceptual demands of each sample.

\bibliography{custom}

\appendix

\section{MathVis-Fine Dataset Statistics}
\label{sec:appendix_stats}

We provide a statistical overview of the MathVis-Fine dataset, which comprises approximately 5,425 high-quality multimodal mathematical problems. To ensure a balanced evaluation of visual necessity, the dataset is categorized into three levels based on the Visual Dependency Score ($\lambda_v$): High ($\lambda_v=1.0$), Medium ($\lambda_v=0.5$), and Low ($\lambda_v=0.0$).

\begin{figure}[h]
    \centering
    \includegraphics[width=\linewidth]{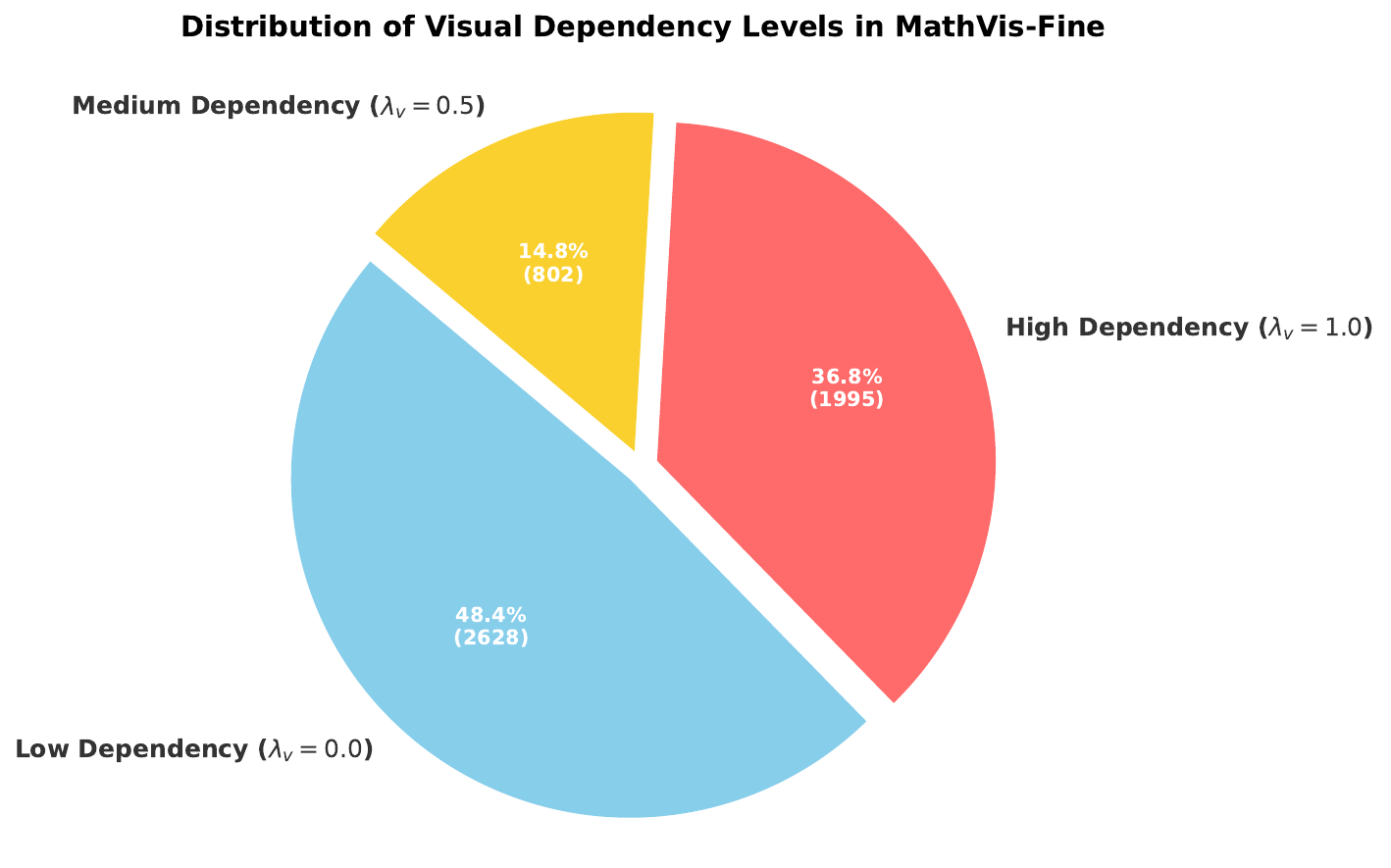} 
   \caption{\textbf{Distribution of Visual Dependency Levels in MathVis-Fine.} The pie chart depicts the percentage of samples annotated as High ($\lambda_v=1.0$), Medium ($\lambda_v=0.5$), and Low ($\lambda_v=0.0$) dependency.}
    \label{fig:data_pie_chart}
    \label{fig:correlation}
\end{figure}

Figure~\ref{fig:data_pie_chart} illustrates the proportional distribution of these three categories. This distribution highlights our strategy to cover a diverse range of multimodal scenarios, from text-dominant problems to those requiring intensive visual interpretation.

\section{Annotation Prompt and Guidelines}
\label{sec:appendix_prompt}

To achieve high-quality and consistent annotations, we employed a standardized prompt to guide the judge model in determining the visual dependency score ($\lambda_v$). The prompt is designed to rigorously evaluate the necessity of visual information for solving the mathematical problem. 

The full system prompt template is provided in Table~\ref{tab:annotation_prompt}.



\begin{table*}[h]
    \centering
    \small
    \begin{tcolorbox}[
        colback=gray!5,      
        colframe=gray!60,    
        title=\textbf{System Prompt for Visual Dependency Annotation},
        fonttitle=\bfseries,
        boxrule=0.8pt,
        arc=2mm
    ]
    
    \textbf{Role Definition:}
    As a multimodal mathematics problem analysis expert, you need to simultaneously analyze two input components: (1) The textual description of the mathematics problem, and (2) The associated image (diagram, geometric figure, statistical graph, etc.). Based on a comprehensive analysis of both, evaluate the degree of visual dependency ($\lambda_v$) required to solve the problem.

    \vspace{0.5em}
    \textbf{Evaluation Criteria:}
    \begin{itemize}[leftmargin=1.5em, itemsep=0.2em]
        \item \textbf{$\lambda_v = 1.0$ (High Visual Dependency / Unsolvable without Image):}
        \begin{itemize}[label=-]
            \item The image contains \textbf{critical mathematical information} absent from the text.
            \item Core data, measurements, coordinate values, angles, or lengths appear \textbf{only in the image}.
            \item Without the image, the problem cannot be solved or can only be guessed.
        \end{itemize}
        
        \item \textbf{$\lambda_v = 0.5$ (Moderate Visual Dependency / Complementary Relationship):}
        \begin{itemize}[label=-]
            \item The text contains the main problem statement, but the image provides \textbf{necessary clarification}.
            \item The image resolves ambiguities (e.g., specific positions of points, relative sizes, label correspondences).
            \item Theoretically solvable from text alone, but the image serves as an authoritative reference that reduces interpretive burden.
        \end{itemize}
        
        \item \textbf{$\lambda_v = 0.0$ (Low Visual Dependency / Decorative Image):}
        \begin{itemize}[label=-]
            \item All necessary mathematical information is \textbf{completely and explicitly} stated in the text.
            \item The image merely replicates textual information or serves as an illustrative example.
            \item Removing the image does not affect problem solvability.
        \end{itemize}
    \end{itemize}

    \vspace{0.5em}
    \textbf{Analytical Framework:}
    \begin{enumerate}[leftmargin=1.5em, itemsep=0.1em]
        \item \textbf{Text Analysis:} Extract explicitly stated variables, conditions, and objectives.
        \item \textbf{Image Analysis:} Identify labels, measurements, and geometric properties in the image.
        \item \textbf{Critical Gap Analysis:} Determine what information in the image is missing from the text and identify any ambiguities the image resolves.
        \item \textbf{Dependency Judgment:} Determine the $\lambda_v$ value based on the analysis.
    \end{enumerate}

    \vspace{0.5em}
    \textbf{Output Specification:}
    Provide ONLY a JSON object with the following structure. No additional text.
    \begin{tcolorbox}[colback=white, colframe=gray!30, boxrule=0.5pt, sharp corners]
    \ttfamily
    \{ \\
    \hspace*{1em} "lambda\_v": <0.0, 0.5, or 1.0>, \\
    \hspace*{1em} "reason": "<Concrete justification stating missing info or clarifications>" \\
    \}
    \end{tcolorbox}
    
    \vspace{0.5em}
    \textbf{Example Assessment Scenarios:}
    \begin{itemize}[leftmargin=1.5em, itemsep=0.1em]
        \item Text: "Find area of shaded region" + Figure with dimensions $\rightarrow$ $\lambda_v=1.0$ (Dimensions only in image).
        \item Text: "In $\triangle ABC$, $AB=5$" + Labeled diagram $\rightarrow$ $\lambda_v=0.5$ (Image clarifies vertex correspondence).
        \item Text: "Solve $x^2-5x+6=0$" + Parabola graph $\rightarrow$ $\lambda_v=0.0$ (Image is decorative).
    \end{itemize}
    
    \end{tcolorbox}
    \caption{The structured system prompt used for automated visual dependency annotation. The prompt guides the judge model to categorize problems into three distinct dependency levels based on the necessity of visual information.}
    \label{tab:annotation_prompt}
\end{table*}

\section{The Use of LLMs}
\label{ap:llm}
In the preparation of this manuscript, we utilized a Large Language Model (LLM). The tool was employed solely for grammar checking and polishing the language expression. All scientific content, analysis, and conclusions remain entirely our own. The authors take full responsibility for the entire content of the paper.

\end{document}